\documentclass[runningheads]{llncs}
\usepackage{graphicx}
\begin{document}
\title{Exploring the Suitability of Semantic Spaces as Word Association Models for the Extraction of Semantic Relationships}
%
%
\author{Epaminondas Kapetanios\inst{1}\orcidID{0000-0002-0617-2183} \and
Vijayan Sugumaran \inst{2}\orcidID{0000-0003-2557-3182} \and
Anastasia Angelopoulou \inst{1}\orcidID{0000-0003-1453-492X}}
\authorrunning{F. Author et al.}
%
\institute{University of Westminster, Regent Street 309, W1B 2HW, London, UK\\
Regent Street 309, W1B 2HW\\
\email{\{kapetae, agelopa\}@westminster.ac.uk}\\
\url{https://www.westminster.ac.uk/research/groups-and-centres/cognitive-computing-research-group} \\
\and
Oakland University, USA\\
\email{sugumara@oakland.edu}}
\maketitle              
\begin{abstract}
Given the recent advances and progress in Natural Language Processing (NLP), extraction of semantic relationships has been at the top of the research agenda in the last few years. This work has been mainly motivated by the fact that building knowledge graphs (KG) and/or bases (KB), as a key ingredient of intelligent applications, is a never-ending challenge, since new  knowledge  needs to be harvested while old knowledge needs to be revised. Currently, approaches towards relation extraction from text are dominated by neural models practicing some sort of distant (weak) supervision in machine learning from large corpora, with or without consulting external knowledge sources. In this paper, we empirically study and explore the potential of a novel idea of using classical semantic spaces and models, e.g., Word Embedding, generated for extracting word association, in conjunction with relation extraction approaches. The goal is to use these word association models to reinforce current relation extraction approaches. We believe that this is a first attempt of this kind and the results of the study should shed some light on the extent to which these word association models can be used as well as the most promising types of relationships to be considered for extraction. 

\end{abstract}
\section{Introduction}

Relationship extraction (RE) is the task of extracting semantic relationships from text. Extracted relationships usually occur between two or more entities of a certain type (e.g. Person, Organisation, Location) and fall into a number of semantic categories (e.g. married to, employed by, lives in). The importance of the outstanding challenges in extracting semantic relationships from text has been exacerbated by the extraction of knowledge bases and/or graphs as the key ingredient of many AI applications (e.g., word meaning disambiguation algorithms, speech recognition, spell checkers). Moreover, work on RE tasks is motivated by the fact that building such knowledge graphs (KG) and/or bases (KB) is a never-ending challenge because, as  the  world  changes,  new  knowledge  needs to be harvested while old knowledge needs to be revised.  

Most work in RE, however, has been witnessed by recent activities show casing progress in natural language processing, such as SemEval series of competitions for relation extractors. In most of these tasks and approaches, RE is performed by forms of prediction of a relationship by looking either at a short span of text within a single sentence containing a single entity pair mention, or spanning over more than one sentence. In any case, the state-of-the-art in RE builds on neural models using distant (a.k.a.  weak) supervision on large-scale corpora for training \cite{ref30}. Despite the various metrics being used, which makes it difficult to compare systems directly, the range of recall has increased over the years as systems improve, with earlier systems having very low precision at 30\% recall. The main metrics used are either precision at N results or plots of the precision-recall. 

Since RE tasks and approaches are very similar with tasks known as Knowledge Base or Graph Embedding (KBE), which are concerned with representing KB entities and relations in a vector space for predicting missing links in the graph, attempts to show that combining predictions from RE and KBE models was beneficial for RE \cite{ref31}. A considerable degree of similarity of RE tasks with classical approaches for the extraction of word associations and meaning, e.g., LSA, LDA, Word2Vec, has been considered as well, since word co-occurrence is the king in these approaches as well.

To the best of our knowledge, however, such latent semantic spaces (LSS) have not been considered as an additional knowledge resource or as a knowledge base to further refine or inform the RE tasks. Combining these two approaches, RE and LSS, could potentially strengthen or weaken the possibility of extracting a semantic relationship, which could be of a more general purpose and not being domain specific.

Having said that, our intention and hope, in this paper, has been to shed some light to the following main question: {\it Is there a potential correlation between the extractors of semantic relations and the latent semantic spaces for topic modelling and word associations?} Answering this question could help us in further pursuing ways to combine these two approaches in order to use RE for knowledge graph/base refinement and updates.

The paper has been organized as follows. Section 2 provides background and related work, specifically about the knowledge base enhanced relation extraction. Section 3 discusses the methodological and experimental approach. Section 4 presents the preliminary results and concluding remarks. 

{\paragraph{\bf Contribution:} The paper explores for the first time the idea and  the possibility of using latent semantic space models to be combined with current approaches for the extraction of semantic relationships in NLP, which are overwhelmingly dominated by neural processing. Using such semantic spaces for word associations, e.g., Word2Vec, could provide good predictions for the kind of semantic relationship we may be looking for.}

\section{Background and related work}

The distinction between RE approaches and those related with the extraction of word associations is not a new one. Generally speaking, since the early 1990s, the line of research around statistical analysis in natural language processing has been split into three main directions:  1) extraction of collocations, which was initiated by Church and Smadja \cite{ref12}, \cite{ref13}, \cite{ref14} and continued by Evert and Krenn \cite{ref18}, Seretan \cite{ref19} and Evert \cite{ref20}, with main applications in translation and language teaching, 2) extraction of word associations and computation of semantic similarities \cite{ref1}, \cite{ref2}, \cite{ref3}, \cite{ref4}, and 3) (semi-)automatic extraction of particular linguistic relations (or thesaurus relations), which are also known as automatic construction of a thesaurus.

The third direction of research is attributed to the (semi-)automatic extraction of particular linguistic relations (or thesaurus relations), e.g., \cite{ref21}, which are also known as automatic construction of a thesaurus. This line of development has been distinguished from the other two lines of research in that it introduces a different methodology based on second order statistics, differentiating between syntactic and paradigmatic relations \cite{ref22}, context comparisons \cite{ref23}. 
Besides, this line of development attempts to give the term ‘word association’ a more precise definition, which can be used to denote various kinds of linguistic relations, often synonyms, sometimes plain word association (play, soccer) and sometimes other linguistic relations like derivation and hyperonymy, antonyms, qualitative direction of adjectives (negative vs. positive). Word sense distinction, contrary to word sense disambiguation, belongs to this area as well, since it describes just another kind of specific relation between words. 

All these approaches, however, do rely on the distributional hypothesis, a mathematically motivated line of influence on today’s computation of relations between words as firstly established by Zelig Harris \cite{ref11}. Another common feature has been that the main goal was to provide information on the general combinatorial possibilities of an entry word. Various types of combinatorial preferences are listed, such as, whether there are any combinatorial preferences of verbs for nouns (e.g. “[to adopt, enact, apply] a regulation”) or what the possible adverbial combinations (i.e. modifications) of a verb are (e.g. “to regret [deeply, very much]”. There is also a distinction between grammatical and lexical collocations with the latter relying on part-of-speech patterns, such as verb-(preposition)-noun, adjective-noun or noun-noun, for permissible collocations in a natural language. For instance, “compose music” and “launch a missile” are permissible, while “compose a missile” is at least awkward.

Generally speaking, associations have been distinguished as {\it association by similarity}, {\it contrast} and {\it contiguity}. Association by similarity is based on the fact that the associated phenomena have some kind of common features. Association by contrast has its origin in what is explained by the presence in phenomena of opposite features. For example, the phenomenon of antonyms: grief - joy, happiness – unhappiness, and so on. Association by contiguity comes into existence when events are situated close together in time or space. Along with them, more complex semantic associations are distinguished. These are, in particular, the association reflecting generic and cause-and-effect relationship between the objects of the world, e.g., a flower - a rose, a disease - death, and so on.

It is also well acknowledged that association is one of the basic mechanisms of memory. In a sense, these mechanisms can be called natural classifiers of the conceptual content of the vocabulary of the language. Ideas and concepts, which are available to the memory are related. This relationship is based on human past experience and, more or less, accurately reproduces an objectively existing relationship between the phenomena of the real world. Under certain conditions, a revival of one idea or concept is accompanied by a revival of other ideas correlated with it. This phenomenon is called the association (a term proposed in the XVIII century by Locke).

In this context, the usage of the term ‘word association’ indicates a broader meaning. In their examples of automatically computed, strongly associated word pairs, there is a mentioning of semantic relations such as meronymy, hyperonymy and so forth. Smadja, however, mentions them as examples of where Church’s algorithm computed just ‘pairs of words’ that frequently appear together’ \cite{ref15}. Lin \cite{ref16} even considers ‘doctors’ and ‘hospitals’ as unrelated and thus wrongly computed as significant by Church and Hanks, although they stand in a meronymy relation.  Nonetheless, other contemporaries, e.g., Dunning \cite{ref17}, improved the mathematical foundation of this research field by introducing the log-likelihood measure. Dunning was the first to coin the term ‘statistical text analysis’.

Despite all these differences, the commonalities between the two worlds, i.e., current RE tasks and approaches, e.g., HRERE \cite{ref32}, and word association extraction models have not been explored further. In HRERE, however, there is only attempt to combine knowledge bases with neural networks targeting relation extraction. A much tighter integration of RE and KBE models is needed with the purpose of using them not only for  prediction,  but  also train them  together,  thus  mutually reinforcing one another. Several other methods have also been proposed \cite{ref33}, \cite{ref34}, \cite{ref35}, \cite{ref36} to use information from KBs to facilitate relation extraction. These vary form considering other relations in the sentential context while predicting  the  target  relation, to utilising  additional side  information from KBs  for  improved  RE. 

In our paper, we take a different approach in that we consider and investigate the possibility, for the first time, to use latent semantic spaces, such as LSA and Word2Vec (pre-trained and no training), for the extraction of word associations as a knowledge base to inform or reinforce relation extraction approaches. Despite the fact that current RE extractors are reporting various metrics, making it difficult to compare systems directly, the main metrics to be used by our approach will be precision at N results or plots of the precision-recall as well and in line with most of the RE approaches.

\section{Methodology and experimentation}

In order to explore the merits of the idea and proposal to use well established semantic spaces for the extraction of word associations for reinforcement of current extractors of semantic relationships, the experimental setup was very much aligned with the SemEval-2010 Task 8, Multi-Way Classification of Semantic Relations between Pairs of Nominals. The task has been, given a sentence and two tagged nominals, to predict the relation between those nominals and the direction of the relation. The dataset contains nine general semantic relations together with a tenth ‘OTHER’ relation.

For instance, given the sentence:

\paragraph{There were apples, pears and oranges in the bowl.}

the semantic relationship\\

\paragraph{(content-container, pears, bowl)}

should be derived.\\

In fact, we used OpenNRE as an open-source and extensible toolkit that provides a unified framework to implement relation extraction models. Subsequently, we applied it to the Google News data set and corpus, in order to extract relations from this corpus. The reason for that is merely the fact that the same corpus has also been used for the extraction of the published pre-trained vectors (e.g., Word2Vec pre-trained) being used for extraction of word associations. This model contains 300-dimensional vectors for 3 million words and phrases. The archive is available at https://code.google.com/archive/p/word2vec/.

Apart from the pre-trained Word2Vec model, we also considered three experimental implementations of LSA, LDA and Word2Vec (no training) from scratch in Python by using Gensim and NumPy, libraries for the first two semantic spaces and the third one, respectively. The intention has been to bring these semantic spaces generated by these classical approaches for the extraction of word associations and meaning into consideration as well.

In order to explore the potential of an interesting correlation between RE tasks and knowledge provided by such semantic spaces, we inputted all the extracted relations as a pair of entities into the four (4) semantic spaces. This would allow us to see whether their association indicated by the RE approach could be reproduced within these semantic spaces. To achieve an objective picture of the relations distribution within the semantic spaces being extracted, we decided to run the experiment with 100 randomly chosen terms underpinned by nouns. 

The results of this empirical study are presented based on the classical precision-recall tandem of metrics as follows: 

\paragraph{{\bf Relation Count (RC):} Number of relations identified within the semantic space. This is an aggregate of all relations being returned/reproduced within the 10 closest terms and for all input terms.}

\paragraph{{\bf Semantic Space Relation Inclusion (SSRIC):} TR / N, N=number of extracted relations from OpeNRE, TR number of input terms, which return at least one relation within the 10 closest.}

\paragraph{{\bf Relation Precision (R-Prec):} Number of correct relations, meaning that these relations have been identified as associations. In other words: R-Prec = Number of all retrieved relations / Number of all correctly identified relations.}

\paragraph{{\bf Relation Recall (R-Rec):} Number of retrieved relations, meaning that these relations have been identified as associations with the input term. In other words: R-Rec = Number of all retrieved relations / Number of all possible relations.}

\section{Experimental results and discussion}

The following table \ref{tab1} summarizes the first results from this empirical study.

\begin{table}
\caption{Precision-recall for reproducing extracted relationships within the semantic similarity spaces for word associations}\label{tab1}
\begin{tabular}{|l|l|l|l|l|}
\hline
  & RC &  SSRIC & R-Prec & R-Rec\\
\hline
LSA &	210	& 0.45	& 0.37	& 0.21\\
LDA &	245	& 0.55	& 0.45	& 0.24\\
Word2Vec	& 235  &	0.65	& 0.48	& 2.4\\
Word2Vec (pre-trained) &	270	& 0.65	& 0.49	& 2.7\\
\hline
\end{tabular}
\end{table}

The following tables shed more light on precision-recall based reproduction of certain categories of extracted relationships within the extracted semantic spaces for word associations.

\begin{table}
\begin{tabular}{|l|l|l|l|l|}
\hline
Cause-Effect Relations	& N / Total	& R-Prec& R-Rec\\
\hline
LSA &	10 / 210	& 0.24	& 0.11\\
LDA &	9 / 245	& 0.35	& 0.12\\
Word2Vec &	7 / 235	& 0.28	& 0.115\\
Word2Vec (pre-trained)	& 13 / 270 & 0.29 &	0.17\\
\hline
\end{tabular}
\end{table}
\begin{table}
\begin{tabular}{|l|l|l|l|l|}
\hline
Component-Whole Relations &	N / Total	& R-Prec &	R-Rec\\
\hline
LSA &	22 / 210 &	0.26 &	0.18\\
LDA &	25 / 245 &	0.36 &	0.19\\
Word2Vec &	26 / 235	& 0.29	& 0.195\\
Word2Vec (pre-trained)	& 28 / 270	& 0.33	& 0.198\\
\hline
\end{tabular}
\end{table}
\begin{table}
\begin{tabular}{|l|l|l|l|l|}
\hline
Content-Container Relations	& N / Total	& R-Prec	& R-Rec\\
\hline
LSA	& 27 / 210	& 0.28	& 0.28\\
LDA	& 24 / 245	& 0.26	& 0.39\\
Word2Vec & 23 / 235	& 0.39	& 0.195\\
Word2Vec (pre-trained) & 26 / 270 &	0.43 &	0.198\\
\hline
\end{tabular}
\end{table}
\begin{table}
\begin{tabular}{|l|l|l|l|l|}
\hline
Entity-Destination	& N / Total	& R-Prec & R-Rec\\
\hline
LSA &	17 / 210	& 0.18	& 0.18\\
LDA &	14 / 245	& 0.16	& 0.19\\
Word2Vec &	13 / 235	& 0.29	& 0.195\\
Word2Vec (pre-trained) & 16 / 270	& 0.23	& 0.198\\
\hline
\end{tabular}
\end{table}
\begin{table}
\begin{tabular}{|l|l|l|l|l|}
\hline
Entity-Origin	& N / Total	& R-Prec & R-Rec\\
\hline
LSA &	25 / 210 &	0.36	& 0.28\\
LDA &	28 / 245 &	0.36	& 0.29\\
Word2Vec & 29 / 235 &	0.39 & 0.24\\
Word2Vec (pre-trained) &	28 / 270 &	0.4	& 0.26\\
\hline
\end{tabular}
\end{table}
\begin{table}
\begin{tabular}{|l|l|l|l|l|}
\hline
Message-Topic &	N / Total	& R-Prec &	R-Rec\\
\hline
LSA &	8 / 210	& 0.28	& 0.38\\
LDA &	25 / 245 &	0.46	& 0.49\\
Word2Vec &	9 / 235 &	0.32 & 0.24\\
Word2Vec (pre-trained)	& 8 / 270	& 0.33 & 0.25\\
\hline
\end{tabular}
\end{table}
\begin{table}
\begin{tabular}{|l|l|l|l|l|}
\hline
Member-Collection &	N / Total	& R-Prec	& R-Rec\\
\hline
LSA &	48 / 210 &	0.48	& 0.35\\
LDA &	45 / 245 & 0.49	& 0.39\\
Word2Vec &	49 / 235	& 0.42	& 0.34\\
Word2Vec (pre-trained)	& 58 / 270	& 0.43	& 0.35\\
\hline
\end{tabular}
\end{table}
\begin{table}
\begin{tabular}{|l|l|l|l|l|}
\hline
Instrument-Agency &	N / Total	& R-Prec	& R-Rec\\
\hline
LSA & 	5 / 210	& 0.21	& 0.28\\
LDA &	10 / 245 &	0.26 & 0.19\\
Word2Vec &	6 / 235	& 0.22 &	0.14\\
Word2Vec (pre-trained)	& 8 / 270	& 0.13	& 0.15\\
\hline
\end{tabular}
\end{table}
\begin{table}
\begin{tabular}{|l|l|l|l|l|}
\hline
Product-Producer &	N / Total	& R-Prec	& R-Rec\\
\hline
LSA &	12 / 210 &	0.31 &	0.35\\
LDA &	15 / 245 &	0.36 &	0.39\\
Word2Vec &	19 / 235 &	0.32	& 0.29\\
Word2Vec (pre-trained) &	18 / 270	& 0.3 &	0.28\\
\hline
\end{tabular}
\end{table}
\begin{table}
\begin{tabular}{|l|l|l|l|l|}
\hline
Other &	N / Total	& R-Prec	& R-Rec\\
LSA &	36 / 210	& 0.16	& 0.18\\
LDA &	50 / 245	& 0.16	& 0.19\\
Word2Vec &	54 / 235 &	0.12 &	0.14\\
Word2Vec (pre-trained)	& 67 / 270	& 0.13	& 0.15\\
\hline
\end{tabular}
\end{table}

\subsection{Discussion of results}

Interesting patterns emerging from this first analysis have been summarized as follows.\\
- Any extracted relations, used as input into the four semantic space models for word associations, do not appear as being among the 10 closest terms for more than 65\% of the input pairs (best case with the Word2Vec pre-trained model).\\
- Following the break down of counts of different extracted relations identified, the highest scores have been encountered for the relation type {\it member-collection}, which is typical for associations of words where hyponymy or hyperonymy (lexical semantics) is hidden. This, in turn, may indicate that these types of relationships are likely to be reproduced within the considered semantic spaces for word associations.\\
- The lowest scores being encountered for the types of relations {\it Instrument-Agency}, {\it product-producer} and {\it Entity-Destination} may denote that these types of relationships can hardly be reproduced within the considered semantic spaces for word associations.\\
- Overall, one may also set up a further hypothesis that many of the detected relations by word association semantic spaces and models, remain undetected by the RE extractors, if they are not combined.

\section{Conclusion}

In this paper, we took a novel idea to combine knowledge extracted in form of word associations with some classical approaches, e.g., LSA, Word2Vec, with current approaches for semantic relationship extraction (RE), to the test. In particular, we conducted an experiment and empirical study to test the hypothesis whether the extracted relations, in the form of pairs of entities or concepts among which a relation has been extracted, returned by current RE approaches can be reproduced within word association semantic spaces. Studying this overlap could shed some light on the potential of a correlation between the two worlds, which have never been used in combination so far. Our aspiration is to inform our RE approach, currently under development, by such word association models and semantic spaces. We believe that certain types of relations, at a more generic level, can be easily detected by combing the two worlds.

%
%
%
%

\end{document}